\documentclass[pdflatex,sn-basic]{sn-jnl}% Basic Springer Nature Reference Style/Chemistry Reference Style
\usepackage{graphicx}%
\usepackage{multirow}%
\usepackage{amsmath,amssymb,amsfonts}%
\usepackage{amsthm}%
\usepackage{mathrsfs}%
\usepackage[title]{appendix}%
\usepackage{xcolor}%
\usepackage{textcomp}%
\usepackage{manyfoot}%
\usepackage{booktabs}%
\usepackage{algorithm}%
\usepackage{algorithmicx}%
\usepackage{algpseudocode}%
\usepackage{listings}%
\theoremstyle{thmstyleone}%
\theoremstyle{thmstyletwo}%

\theoremstyle{thmstylethree}%

\begin{document}

\title[Explainable fault and severity classification for rolling element bearings using Kolmogorov-Arnold networks]{Explainable fault and severity classification for rolling element bearings using Kolmogorov-Arnold networks}


\author*[1]{\fnm{Spyros} \sur{Rigas}}\email{spyrigas@uoa.gr}

\author[2]{\fnm{Michalis} \sur{Papachristou}}\email{mixpap@phys.uoa.gr}

\author[3]{\fnm{Ioannis} \sur{Sotiropoulos}}\email{jsotiropoulos@mail.ntua.gr}

\author[1]{\fnm{Georgios} \sur{Alexandridis}}\email{gealexandri@uoa.gr}

\affil[1]{\orgdiv{Department of Digital Industry Technologies}, \orgname{School of Science, National and Kapodistrian University of Athens}, \city{Psachna}, \postcode{34400}, \country{Greece}}

\affil[2]{\orgdiv{Department of Physics}, \orgname{School of Science, National and Kapodistrian University of Athens}, \city{Athens}, \postcode{15784}, \country{Greece}}

\affil[3]{\orgname{School of Electrical \& Computer Engineering, National Technical University of Athens}, \city{Athens}, \postcode{15780},  \country{Greece}}


\abstract{Rolling element bearings are critical components of rotating machinery, with their performance directly influencing the efficiency and reliability of industrial systems. At the same time, bearing faults are a leading cause of machinery failures, often resulting in costly downtime, reduced productivity, and, in extreme cases, catastrophic damage. This study presents a methodology that utilizes Kolmogorov-Arnold Networks to address these challenges through automatic feature selection, hyperparameter tuning and interpretable fault analysis within a unified framework. By training shallow network architectures and minimizing the number of selected features, the framework produces lightweight models that deliver explainable results through feature attribution and symbolic representations of their activation functions. Validated on two widely recognized datasets for bearing fault diagnosis, the framework achieved perfect F1-Scores for fault detection and high performance in fault and severity classification tasks, including 100\% F1-Scores in most cases. Notably, it demonstrated adaptability by handling diverse fault types, such as imbalance and misalignment, within the same dataset. The symbolic representations enhanced model interpretability, while feature attribution offered insights into the optimal feature types or signals for each studied task. These results highlight the framework's potential for practical applications, such as real-time machinery monitoring, and for scientific research requiring efficient and explainable models.}




\maketitle

\section{Introduction}\label{sec1}

Rotating machinery plays an indispensable role in modern industry, powering numerous applications across the manufacturing, energy, and transportation sectors \citep{IEEEVibration}. Among their components, rolling element bearings (referred to simply as bearings hereafter) are vital, yet vulnerable elements, with a heavy influence on the performance and lifespan of machines. Notably, it has been reported that up to 50\% of motor faults are bearing-related \citep{IEEEMotorsReview}, while several issues in rotating machinery today can be traced to the improper design or application of bearings \citep{ElsevierBearingsReview}. Such failures can result in severe consequences, including unexpected downtime and costly repairs \citep{RotMachReview,JIMSImbalanced}, or even catastrophic damage or loss of life in extreme cases \citep{CWRUReview}. Moreover, in manufacturing environments, where continuous production is critical, disruptions caused by bearing failures can lead to substantial losses in productivity \citep{GABoT}. Early and accurate bearing fault detection and classification are therefore essential in modern industrial and manufacturing practices.

Before the widespread use of machine learning (ML) and deep learning (DL) methodologies, bearing fault detection and classification relied on other widely adopted techniques to identify characteristic fault patterns. For instance, vibration analysis was commonly employed to detect frequency peaks associated with specific faults \citep{IEEEMotorsReview}, while signatures of these vibration frequencies were also identified in the current spectrum through electrical signal processing \citep{ElectrPeaks}. Additionally, features extracted from the Fourier spectrum of vibration signals were utilized to detect faults as peaks in the frequency-domain \citep{SpectralKurtReview} and acoustic emission (AE) monitoring offered earlier and more fine-grained fault detection compared to vibration monitoring \citep{AEMethod}. However, these methods were often bound to specific fault scenarios or experimental conditions, which limited their applicability to diverse operating environments. For example, directly identifying bearing faults through raw vibration signals is challenging, as vibrations are typically dominated by imbalance and misalignment components \citep{BasicVibGuide}. Moreover, the experimental results of \cite{ElectrPeaks} were based on cases of extensive bearing damage, raising concerns about the applicability of this approach for less severe faults \citep{BasicVibGuide}. Finally, features like spectral kurtosis have been shown to be sensitive to strong harmonic interferences when used as the only fault indicator \citep{AdaptSpectrKurt} and AE monitoring has also proved to be highly susceptible to background noise \citep{AEVibComp}.

Despite their limitations, the aforementioned techniques laid the foundation for identifying which types of sensor data are most effective for detecting and classifying bearing faults. Modern data-driven approaches have built upon this groundwork, incorporating features extracted from sensor data to develop more robust and generalizable frameworks. Examples of features extracted directly from the time-domain signal include but are not limited to the root mean square (RMS), crest factor (CF), skewness and kurtosis  \citep{StatisticalFeatures}. Spectral features, such as fundamental frequencies, spectral kurtosis and spectral entropy are obtained by applying a Fast Fourier Transform (FFT) to the time-domain signals and have also been widely used in such applications \citep{FullFeatureReview}. Beyond time- and frequency-domain features, time-frequency representations, such as those derived from the Short-Time Fourier Transform or wavelet transformations, are often used to extract features for capturing transient and non-stationary behaviors \citep{TimeFreqRepresentations}. Drawing from these diverse feature sets, a series of ML models such as $k$-Nearest Neighbors \citep{kNN1,kNN2}, Support Vector Machines (SVMs) \citep{SVM1,SVM2}, and Random Forests \citep{RF1,RF2,RF3} have been explored for the tasks of bearing fault detection and classification. More recently, the widespread adoption of sensors in industrial settings and the advent of the big data era have driven the use of DL architectures for these tasks, including Autoencoders \citep{AutoEncoder,RNN1}, Recurrent Neural Networks \citep{RNN1,RNN2}, Convolutional Neural Networks \citep{CNN1,CNN2,CNN3}, and Generative Adversarial Networks \citep{GAN1,GAN2}.%, to name but a few.

Albeit successful in achieving high performance for bearing fault detection and classification, ML and especially DL models often fall short in areas where traditional approaches excel, with explainability being a notable example. The ability to understand and interpret a model's decisions is crucial, especially in safety-critical applications or when deeper insights into the underlying physical processes are required \citep{ExplainabilityRequired}. In addition to explainability, a significant challenge arises in deploying DL models for real-time condition-based monitoring on edge devices, i.e., resource-constrained computing units located close to the machinery they monitor. Many DL architectures are computationally intensive, making them unsuitable for resource-constrained environments \citep{HighComplexity}. Another important consideration is the quantity and quality of features used by these models. While leveraging a large number of features can often yield superior results, achieving comparable performance with fewer features is far more desirable \citep{LargeFeatureNumber}; budget constraints and technical limitations in practical scenarios demand careful sensor selection, as collecting an exhaustive set of measurements is neither feasible nor economical. Furthermore, the effectiveness of features can vary significantly, depending on the dataset or system under study; features that perform well for one problem may be suboptimal for another. This variability highlights the need for models capable of adaptively selecting the most relevant features for a given problem \citep{FeatureEffectiveness}. To address these challenges, this paper presents a unified framework centered around Kolmogorov-Arnold Networks (KANs), designed to provide explainability, efficiency, and adaptive feature selection.

Inspired by the Kolmogorov-Arnold representation theorem, KANs were recently introduced by \cite{KAN1} as an alternative to Multi-Layer Perceptrons (MLPs), serving as a new paradigm for the underlying architecture of DL models. Unlike in the case of MLPs, where activation functions are fixed, KANs contain trainable univariate functions as activations, allowing them to represent relationships in symbolic forms. This inherent explainability, along with their demonstrated performance in domains such as differential equations \citep{PDE1,PDE2,PDE3}, high-energy physics \citep{HEP1,HEP2}, and smart systems and devices \citep{SmartKAN1,SmartKAN2}, makes KANs a promising candidate for addressing both scientific and engineering problems \citep{KAN2}. In the context of bearing fault detection and classification, there is a notable lack of studies utilizing KANs, with the exception of \cite{NaturalGas}. In that study, the Case Western Reserve University (CWRU) bearing dataset \citep{CWRU} was employed; however, the primary focus of the paper was unrelated to bearing fault diagnosis and instead aimed at the early prediction of natural gas pipeline leaks.

Building on the potential of KANs and addressing the identified gaps, the main aspects of the proposed framework - and thus the main contributions of the current work - can be summarized in the following points:

\begin{itemize}
	\item \emph{Explainable selection of minimum features}: by training shallow KANs with sparsity-inducing regularization, the minimum number of features relevant to the problem can be automatically identified via attribution scores and dynamic thresholds.
	\item \emph{Interpretable and lightweight model design}: a model in symbolic form can be obtained from the trained activation functions, enabling its analysis outside of a black-box regime and ensuring efficiency for deployment on edge devices.
	\item \emph{A unified approach to bearing fault diagnosis}: fault detection \& classification, and severity classification are addressed within the same framework.
	\item \emph{Broad applicability beyond bearing faults}: the framework's generalization capabilities are demonstrated by its application to non-bearing-exclusive data.
\end{itemize}

\noindent To the best of the authors' knowledge, this is among the first attempts to address bearing faults in a holistic and generalizable manner, incorporating detection, classification and severity estimation within a single, lightweight DL framework that also handles feature selection and provides explainable results. A recent related study by \cite{GABoT} touches upon some of these challenges by employing an SVM model for fault classification and genetic algorithms (GAs) for automated feature selection. Nevertheless, GAs are computationally demanding, and SVMs, along with GAs, lack the explainability provided by the proposed KAN-based approach.

The remainder of the present paper is structured as follows: Section \ref{sec2} presents the proposed framework in detail, including its components, methodology and theoretical foundation. Subsequently, the two datasets utilized in this study are introduced in Section \ref{sec3}, along with a discussion on the rationale for their selection and a presentation of the feature libraries extracted to implement the framework. In Section \ref{sec4} the experimental results obtained on both datasets are reported, focusing on selected features, model performance and symbolic representations. Finally, Section \ref{sec5} provides a summary and discussion on the work's main findings.

\section{Proposed Methodology}\label{sec2}

Prior to the discussion of the proposed methodology's technical details, an overview of KANs, their theoretical formulation and the properties that establish them as a key component of the framework are provided.

\subsection{Kolmogorov-Arnold Networks}\label{sec2.1}

The theoretical foundations of KANs lie on the Kolmogorov Superposition Theorem (KST), which provides a robust theoretical framework for decomposing multivariate functions into simpler univariate functions through summation operations. Earlier attempts to apply the theorem for function approximation sought to implement it in its exact form but faced significant challenges due to the pathological behavior of the inner univariate functions \citep{Perdikaris}. Recently, \cite{KAN1} extended the KST into a ``deep'' equivalent, introducing a more flexible network architecture that contains an arbitrary number of layers with arbitrary widths, rather than adhering strictly to the original formulation of the theorem.

%\subsubsection{Architecture}\label{sec2.1.1}

One such extended architecture with $L$ layers is defined by an integer array $\left[n_0, n_1, ..., n_L\right]$, where $n_i$ denotes the number of input nodes of the $i$-th layer. Unlike in MLPs, a KAN layer corresponds to the activation functions between a set of input and output nodes, rather than the inputs or outputs themselves, which is why the array has $L+1$ elements. The corresponding model can be written as

\begin{equation}
	u\left(x;\theta\right) = \left[\Phi^{(L)}\circ \dots \circ \Phi^{(1)}\right]\left(x\right), \label{eq1}
\end{equation}

\noindent where $\theta$ represents the network's trainable parameters, $\circ$ denotes successive application of $\Phi^{(l)}$, and

\begin{equation}
	\Phi^{(l)}\left(x^{(l)}\right) = 	
	\begin{pmatrix}
		\phi_{l,1,1}(\cdot) & \cdots & \phi_{l,n_l,1}(\cdot) \\
		\phi_{l,1,2}(\cdot) & \cdots & \phi_{l,n_l,2}(\cdot) \\
		\vdots & \ddots & \vdots \\
		\phi_{l,1,n_{l+1}}(\cdot) & \cdots & \phi_{l,n_{l},n_{l+1}}(\cdot)
	\end{pmatrix} x^{(l)}, \label{eq2}
\end{equation}

\noindent with $\phi_{l,i,j}$ being the $l$-th layer's activation function, which connects the layer's $i$-th input node to its $j$-th output node. These activation functions are given by

\begin{equation}
	\phi\left(x\right) = c_rr\left(x\right) + c_BB\left(x\right), \label{eq3}
\end{equation}

\noindent where

\begin{equation}
	r\left(x\right) = \frac{x}{1+ \exp\left(-x\right)} \label{eq4}
\end{equation}

\noindent is the Sigmoid Linear Unit (SiLU) function and

\begin{equation}
	B\left(x\right) = \sum_{i=1}^{G+k}{c_i B_i\left(x\right)} \label{eq5}
\end{equation}

\noindent is a spline activation composed of $\left(G+k\right)$ B-spline basis functions of order $k$ on a grid with $G$ intervals. The parameters $c_r$, $c_B$ and $\left\{c_i\right\}_{i=1}^{G+k}$ of each activation function, and consequently the activation function itself, are trainable, which is where the interpretability of KANs stems from: each trained activation function can be replaced with a symbolic representation that best fits it.

Beyond this interpretability, KANs feature an attribution scoring mechanism that provides explainability by quantifying the relative importance of the model's input features \citep{KAN2}. This requires defining the attribution score of the $i$-th input node in the $l$-th layer, denoted by $A_{l,i}$. To compute such scores, one must also define the standard deviation of the $l$-th layer's activation function connecting node $i$ to node $j$ as

\begin{equation}
	E_{l,i,j} = \sqrt{\frac{1}{N}\sum_{s=1}^{N}{\left[\phi_{l,i,j}\left(x_s\right)- \frac{1}{N}\sum_{p=1}^{N}{\phi_{l,i,j}\left(x_p\right)}\right]^2}}, \label{eq6}
\end{equation}

\begin{figure}[b]
\centering
\includegraphics[width=0.55\textwidth]{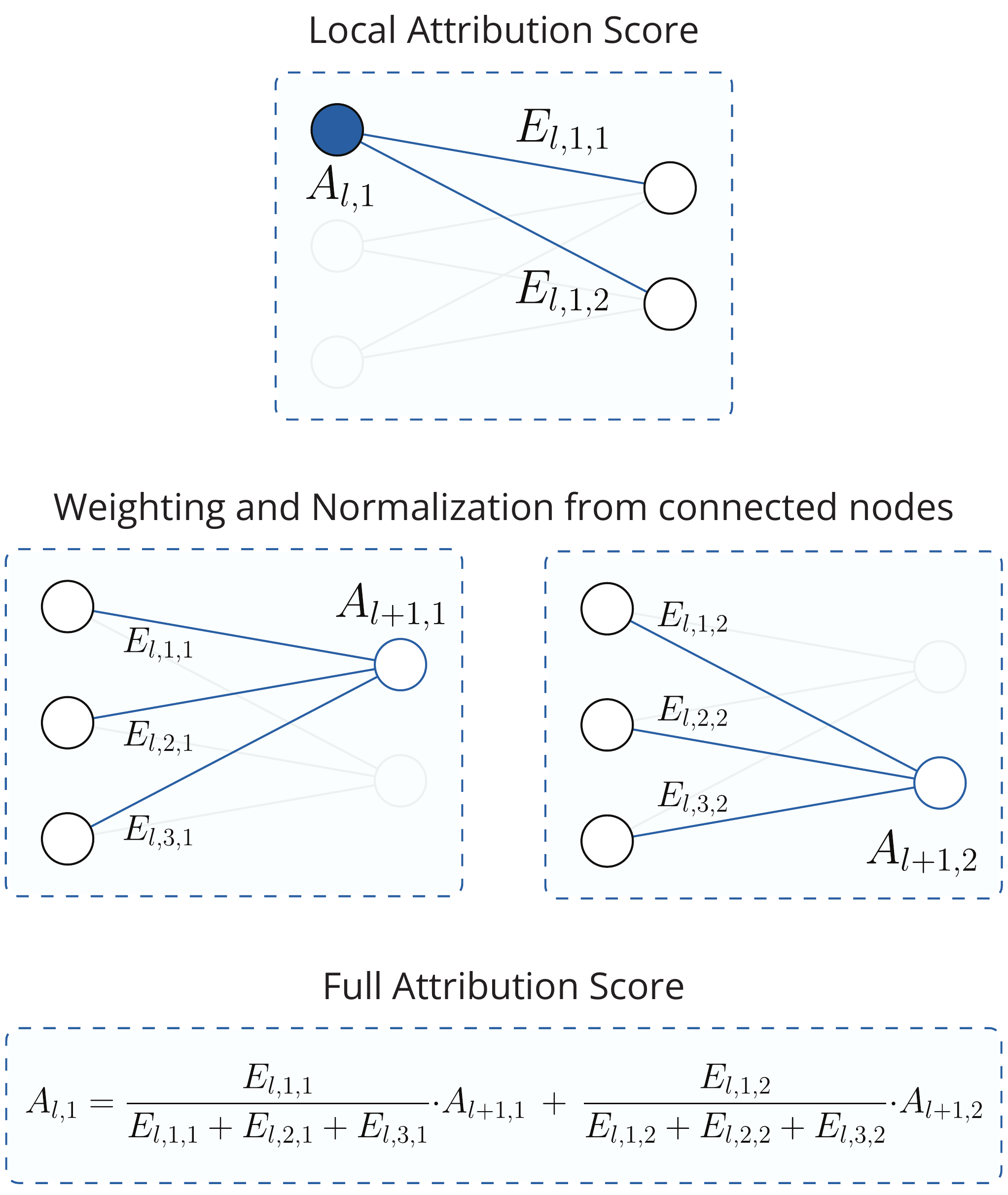}
\caption{Example of an iteration of node score attribution in a layer of 3 input and 2 output nodes.}
\label{attribution}
\end{figure}

\noindent where $N$ is the number of samples. Then, $A_{l,i}$ can be calculated recursively via

\begin{equation}
	A_{l,i} = \sum_{j=1}^{n_{l+1}}{E_{l,i,j}\cdot A_{l+1,j}\cdot\left(\sum_{p=1}^{n_l}{E_{l,p,j}}\right)^{-1}}, \hspace{5mm} l \in \left\{L, \dots, 1\right\}, \label{eq7}
\end{equation}

\noindent and the initial condition $A_{L+1,i} = 1, \forall i \in \left\{1, ..., n_L\right\}$, i.e. setting the scores of the final layer's output nodes to 1. Note that the notation refers to input nodes, which is why the subscript is $L+1$, even though the model consists of $L$ layers. The final score for the $i$-th input feature then simply corresponds to $A_{1,i}$.

This attribution score captures the global importance of a node, as it accounts for more than just the activations on its outbound edges via $E_{l,i,j}$, which are highly local. Instead, $E_{l,i,j}$ is adjusted by the attribution scores $A_{l+1,j}$ of all subsequent nodes, $j$, that are linked to node $i$. Additionally, for each of these subsequent nodes, the expression is also normalized by $\sum_{p=1}^{n_l}{E_{l,p,j}}$, which takes into account all activations on that node's inbound edges. Essentially, the recursive relation in Eq. \eqref{eq7} ensures that the importance of the $i$-th node in the $l$-th layer reflects that node's contributions on the entire network downstream from the node, underscoring its global nature. The attribution scoring mechanism is illustrated for an example architecture in Fig. \ref{attribution}, which provides a visual representation of how each node's score is influenced by subsequent nodes and the activations along their connecting edges.

\subsection{Proposed Framework}\label{sec2.2}

Building on these theoretical foundations, the initial objective of the proposed methodology is to leverage the attribution scoring mechanism of KANs for automatic feature selection. To this end, a feature library is first constructed using existing literature and domain knowledge, automatic feature extraction methods (e.g., \cite{AutoFeatureExtr}), or a combination thereof, depending on the studied problem (bearing faults in the present case). Each data sample is thus represented as a $K$-dimensional vector, where $K$ is equal to the total number of features in the library. After all data samples are split into distinct training, validation, and evaluation sets, the feature selection process is formulated as a grid-search multi-objective problem.

Specifically, multiple KAN model instances are trained on the training set by minimizing a loss function that incorporates the following regularization term:

\begin{equation}
	\mathcal{L}_\text{reg} = \lambda\sum_{l=1}^{L}\left[\mathbf{A}_l - \sum_{i=1}^{n_l}{\frac{A_{l,i}}{\mathbf{A}_l} \log\left(\frac{A_{l,i}}{\mathbf{A}_l}\right)} \right], \label{eq8}
\end{equation}

\noindent where

\begin{equation}
	\mathbf{A}_l = \sum_{i=1}^{n_l}{A_{l,i}}. \label{eq9}
\end{equation}

\noindent This sparsity-inducing expression corresponds to a mixture of L1 and Entropy regularization (first and second summand of Eq. \eqref{eq8}, respectively), with an overall weight $\lambda$, corresponding to a hyperparameter. Each model instance is trained for a fixed number of epochs using a distinct value of $\lambda$, selected from a discretized range $\left[\lambda_\text{min}, \lambda_\text{max}\right]$. For each model instance corresponding to a particular $\lambda$, the attribution score is computed for all $K$ features using Eq. \eqref{eq7}. The most important features are then selected based on the condition

\begin{equation}
	A_{1,i} \geq \tau, \label{eq10}
\end{equation}

\noindent where $\tau$ is a hyperparameter which determines the threshold for feature selection. To ensure an appropriate choice of $\tau$, multiple threshold values are evaluated, drawn from a discretized range $\left[\tau_\text{min}, \tau_\text{max}\right]$, to identify the most significant features. As a result, each combination of $\left(\lambda, \tau\right)$ corresponds to a distinct selection of important features.

The optimal values of $\lambda$ and $\tau$ are determined based on a multi-objective criterion: maximizing the performance of a trained model on the validation set while minimizing the number of selected features. To achieve this, model instances are retrained, this time without regularization ($\lambda = 0$), and only using the features selected for each combination of $\lambda$ and $\tau$. The results are then analyzed to identify the Pareto front \citep{pareto}, representing the trade-off between model performance and feature count. If the Pareto front contains a single element, the corresponding $\left(\lambda, \tau\right)$ pair is deemed the optimal. In cases where multiple points lie on the Pareto front, an additional rule can be applied to make the final selection - for instance, choosing parameters which led to the model that achieves the highest performance while using up to $k$ features, where typically $k \ll K$.

With the feature selection phase complete, the methodology transitions to the model selection phase, which involves hyperparameter tuning. This stage focuses on KAN-specific parameters, namely the grid size, $G$, and the grid adaptability factor, $0 \leq g_e \leq 1$, defined via

\begin{equation}
	\mathcal{G} = g_e \mathcal{G}_u + \left(1-g_e\right)\mathcal{G}_a, \label{eq11}
\end{equation}

\noindent where $\mathcal{G}_u$ and $\mathcal{G}_a$ correspond to purely uniform and purely adaptive grids, respectively, and $\mathcal{G}$ is a linear combination of the two \citep{KAN1}. During the feature selection phase, the model instances are trained using a fixed, small grid to prioritize computational efficiency. In the model selection phase, larger grid sizes are also considered, in order to allow the trainable activation functions to capture more intricate patterns in the data by using a greater number of B-spline basis functions. Additionally, incorporating grid adaptivity has been shown to enhance model performance by adapting each layer's grid to the underlying data structure of its inputs \citep{KAN1,AdaptiveKAN}. Hyperparameter tuning is conducted similarly to the feature selection process: a KAN model instance is trained for each configuration of $\left(G, g_e\right)$ over a fixed number of epochs, and its performance is evaluated on the validation set. However, in this phase, performance is assessed along two axes: the performance of the trained KAN itself, as well as the performance of its symbolic version.

As discussed in the context of KANs' interpretability, the trained activation functions can be replaced with symbolic functions selected from a predefined library. The optimal symbolic function for each layer is determined as the one minimizing a cost function that balances high $R^2$ values of the symbolic fit against the symbolic function's complexity. Complexity can be assigned in various ways, for example based on the statistical occurrence of functions in physical formulae \citep{PhysicsEqs}. In this work, complexities are assigned in a way that ensures consistency with the PySR framework \citep{PySR}, widely recognized as the state-of-the-art open-source solution for symbolic regression tasks. In the present work, the cost function is given by

\begin{equation}
	\mathcal{C}\left(C,R^2\right) = \exp\left(\alpha C\right) + \beta\ln\left(1-R^2\right), \label{eq12}
\end{equation}

\noindent \noindent where $R^2$ is the fit's $R^2$ score, $C$ denotes the assigned complexity of the symbolic function, and $\alpha$ and $\beta$ are parameters that control the penalty for function complexity and the reward associated with the fit quality, respectively. The symbolic library utilized in this work is provided in Appendix \ref{appA}, along with all assigned complexities. Naturally, such libraries cannot be exhaustive, but they provide a good starting point for deriving interpretable symbolic representations.

\begin{figure*}[t]
	\centering
	\includegraphics[width=\textwidth]{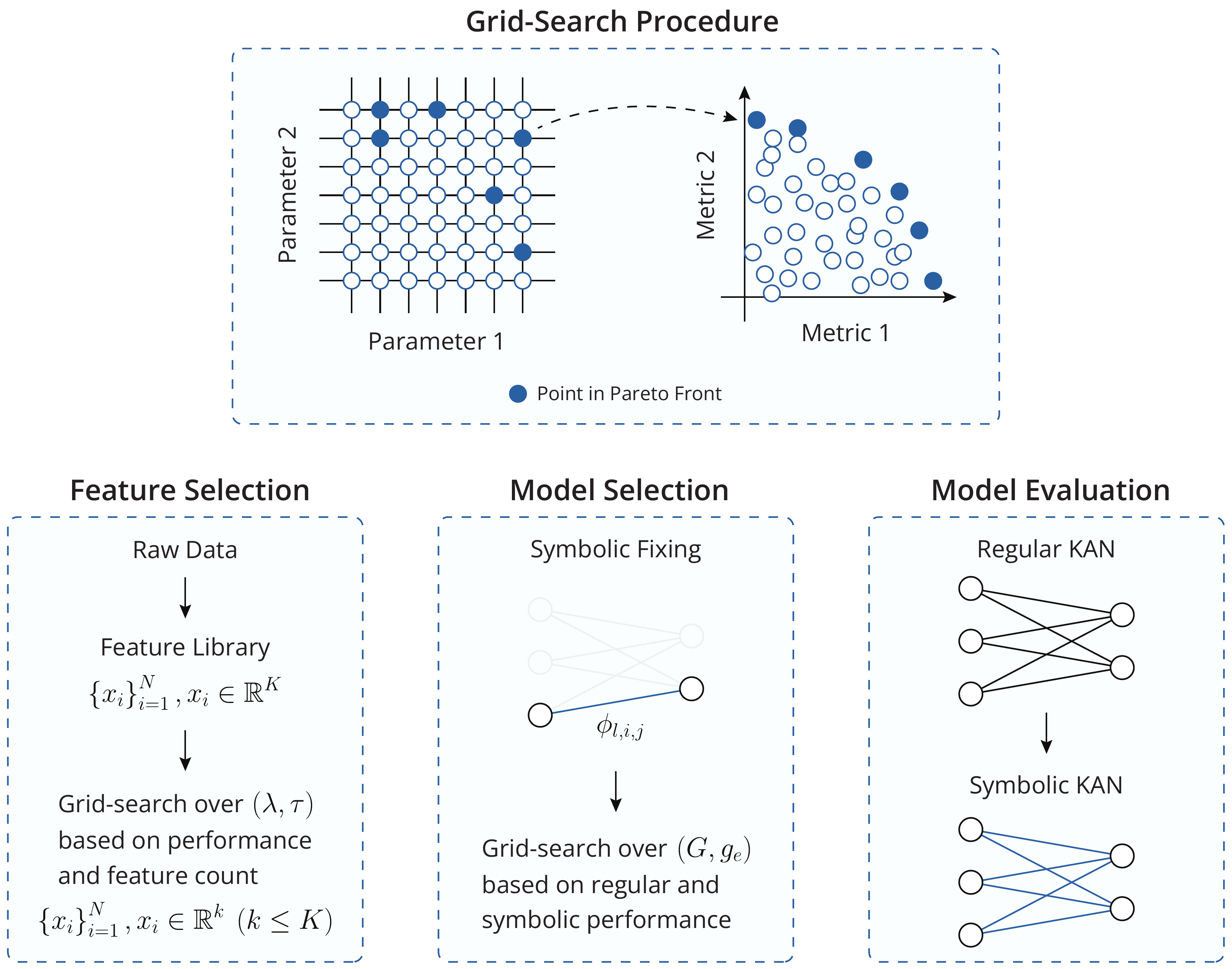}
	\caption{Schematic representation of the proposed framework, encompassing the feature selection, model selection and model evaluation processes.}
	\label{framework}
\end{figure*}

Following each trained model's evaluation on the validation set, both with and without symbolic fitting, the Pareto front for the performance metrics of the regular and symbolic KAN models is identified. Similar to the feature selection phase, if the Pareto front is a singleton, then the corresponding $\left(G, g_e\right)$ values are selected. However, if the Pareto front includes multiple points, additional rules are necessary for the selection. With the most important features identified during the feature selection phase and the optimal model hyperparameters determined during model selection, the training and validation sets are combined, and a final model instance is trained on their union. This model is subsequently evaluated on the isolated evaluation set to provide a final and unbiased performance assessment. A symbolic version of this final model is also extracted. While its performance is generally expected to be lower than that of the regular KAN model, the symbolic version offers a more interpretable alternative, which may be preferable in scenarios where explainability is prioritized over optimal performance. A schematic representation of the complete framework can be seen in Fig. \ref{framework}.

\section{Datasets and Feature Extraction}\label{sec3}

As previously outlined, the proposed framework has been designed for applicability across a wide range of problems beyond bearing faults. To apply it for bearing fault detection and classification, two widely recognized datasets are selected: the CWRU bearing dataset \citep{CWRU} and the Machinery Fault Database (MaFaulDa) dataset \citep{MAFAULDA,MaFaulDaPaper}. The CWRU dataset is chosen due to its characterization as a dataset where feature selection is highly nontrivial, containing data that deviate from the typical characteristics expected for certain fault types \citep{CWRUReview}. The MaFaulDa dataset, on the other hand, is selected for its broader scope, as it includes not only bearing faults but also additional types of machinery faults, thereby enabling the demonstration of the framework's generalizability within a single dataset. Before detailing the process of constructing a feature library from the raw time-series signals of the two datasets, a more detailed introduction to each dataset is provided Sections \ref{sec3.1} and \ref{sec3.2}.

\subsection{CWRU Dataset}\label{sec3.1}

The CWRU dataset was generated using a test rig designed to simulate bearing faults under controlled conditions. The setup consisted of a 2-horsepower motor, a torque transducer, and a dynamometer, with the test bearings supporting the motor shaft. Three types of single-point faults were induced in the bearings using electro-discharge machining: inner raceway (IR), ball (B), and outer raceway (OR) faults, with fault diameters ranging from 7 mils (1 mil is equivalent to 0.001 inches) to 40 mils. Faults were applied to both the drive-end and fan-end bearings. The dataset comprises vibration measurements collected using accelerometers attached to the motor housing at the 12 o'clock position for both the drive-end and fan-end bearings, with an additional accelerometer attached to the base plate in some experiments. The signals were recorded at sampling rates of 12 kHz and, for certain drive-end faults, 48 kHz. For OR faults, experiments were conducted at different positions relative to the load zone (3 o'clock, 6 o'clock, and 12 o'clock) to capture variations in the vibration response. Thus, the dataset contains six classes for classification, labeled as N (normal), B, IR, OR@3, OR@6, and OR@12.

The original dataset's files can be categorized along several axes. Based on motor speed, the files are divided into four groups: 1730, 1750, 1772, and 1797 rotations per minute (RPM). Based on fault location and sampling rate, the dataset includes normal files measured at 48 kHz, drive-end faults measured at 12 kHz, fan-end faults measured at 12 kHz, and drive-end faults measured at 48 kHz. Additionally, the files differ in the time-series data they contain: some include only drive-end measurements, most include both drive-end and fan-end measurements, and a few include drive-end, fan-end, and base measurements. Due to the inconsistencies mentioned in \cite{MarineCWRU}, the version of the dataset curated for the purposes of the cited work was used. Moreover, all 48 kHz drive-end measurements were excluded for two reasons: to ensure uniformity, as corresponding fan-end measurements at 48 kHz are unavailable, and because these files exhibited significant variability in the number of measurements, making them unsuitable for consistent sampling for the purposes of feature extraction. This process resulted in a total of 101 retained files.

\subsection{MaFaulDa Dataset}\label{sec3.2}

The MaFaulDa dataset was created using a test rig designed to emulate the dynamics of motors with two shaft-supporting bearings. It comprises multivariate time-series data collected from sensors mounted on a SpectraQuest alignment/balance vibration trainer machinery fault simulator. The sensors included one triaxial accelerometer for the underhang bearing (bearing located between the motor and rotor) and three industrial accelerometers for the overhang bearing (bearing located outside the rotor, opposite the motor), oriented along the axial, radial, and tangential directions. Additionally, an analog tachometer measured the system's rotational frequency, and a microphone captured operational sound. All signals were recorded at a sampling rate of 50 kHz over a duration of 5 seconds.

The dataset includes scenarios representing both normal operation and various fault conditions. In the normal class (N), the system operated without faults across 49 distinct rotation frequencies, ranging from 737 to 3686 RPM in approximately 60 RPM intervals. Bearing faults, similar to those in the CWRU dataset, involved defects in the inner raceway (IR), ball (B), and outer raceway (OR). These faults were studied in both bearings, underhang and overhang, one at a time. To ensure fault detectability, additional imbalances of 6 g, 10 g, and 20 g were introduced. Bearing fault scenarios were recorded under 49 rotation frequencies for lighter imbalances, while fewer frequencies were studied for heavier ones due to increased vibrations.

Beyond bearing faults, the dataset also includes additional machinery faults, namely imbalance (I) and axis misalignment. Imbalance faults were simulated by attaching varying load weights (6 g to 35 g) to the rotor. For weights up to 25 g, all 49 rotation frequencies were studied, whereas higher weights limited the maximum frequency to 3300 RPM due to increased vibrations. Axis misalignment was divided into horizontal misalignment (HM) and vertical misalignment (VM), induced by shifting the motor shaft by offsets of 0.5 mm to 2.0 mm for the former, and 0.51 mm to 1.9 mm for the latter. For each misalignment severity, the same 49 rotation frequencies as in the normal class were studied. In total, the dataset corresponds to 10 distinct classes and comprises 1951 data files, all of which were retained for feature extraction.

\subsection{Feature Library}\label{sec3.3}

The extracted features were acquired by first augmenting and then preprocessing data from both datasets. Data augmentation was particularly critical for the CWRU dataset, which contained only four data files per fault type and severity - corresponding to the four rotational frequencies studied. In contrast, the MaFaulDa dataset included nearly 50 examples per fault case, yet augmentation was still applied to further enhance the dataset. The first step involved identifying the rotational frequency, $f_r$, for each file. For the CWRU files, the exact RPM values were already known. However, for the MaFaulDa dataset, where RPM values were estimated per file, the rotational frequency was calculated using the two-step algorithm proposed in \cite{FRCalc}, based on the tachometer signal. This method was selected to avoid misidentification of $f_r$, which could otherwise be obscured by spectral peaks introduced by machine faults in the signal's frequency spectrum.

Once the rotational frequency was determined, it was combined with the sampling rate, $F_s$, to split each time-series into smaller segments of $N\cdot F_s/f_r$ data points, where $N$ represents the number of complete motor rotation cycles. The choice of $N$ balances a trade-off between dataset size and segment quality: a smaller value leads to more segments but at the cost of lower quality, while a larger value preserves the original time-series' quality at the expense of limited samples. For this study, $N=48$ was chosen as a compromise, yielding approximately six segments per file for the CWRU dataset. The same number of cycles was chosen for the MaFaulDa dataset to maintain consistency, resulting in augmented datasets containing 603 and 6268 segments for CWRU and MaFaulDa, respectively.

Using the augmented datasets, a series of time-domain, frequency-domain, and time-frequency features were extracted, based on established literature in machinery fault diagnosis. For the time-domain features, the extracted metrics included the RMS, mean, variance, skewness, kurtosis, entropy, shape factor, crest factor, impulse factor, and margin factor, along with histogram upper and lower bounds as described in \cite{FullFeatureReview}. From the frequency domain, spectral skewness and kurtosis were calculated after applying an FFT to each signal. Additionally, the signal magnitudes at the fundamental frequency and its first two harmonics were extracted, following \cite{MaFaulDaPaper}.

For time-frequency features, wavelet transformations were employed, as they are highly effective for identifying the machinery faults. The \texttt{pywavelets} library \citep{PyWT} was used to perform a multilevel decomposition of order 4 on each segment, utilizing a biorthogonal wavelet. Following \cite{GABoT}, features derived from the fine-grained wavelet coefficients included the mean, median, RMS, standard deviation, variance, skewness, kurtosis, and entropy. Percentile values at the 5th, 25th, 75th, and 95th levels were also extracted, along with the number of mean and zero crossings. Using this approach, a feature library of 62 and 243 features was compiled for the augmented CWRU and MaFaulDa datasets, respectively. This corresponds to 31 features per dataset signal, with the exception of the tachometer signal in MaFaulDa, from which no spectral features were extracted \citep{MaFaulDaPaper}.

A detailed list of all extracted features for this work is provided in Appendix \ref{appB}. It should be noted that certain features overlap in terms of the information they encode; for instance, the impulse factor is the product of the crest and shape factors, while the standard deviation is the square root of the variance. This intentional redundancy allows the framework to identify the most relevant features automatically and discard the rest during the feature selection process. After all, if the most effective features for the studied problem were already known, the feature selection phase would be redundant.

\section{Experimental Results}\label{sec4}

This section presents the experimental findings obtained by applying the proposed framework to the two constructed feature libraries, addressing three distinct tasks: fault detection, fault classification, and severity classification for each fault type. Exclusively shallow KANs, i.e., models with a single layer, were considered throughout the experiments to minimize the number of model parameters, thus ensuring that the models remain lightweight and their symbolic representation does not become overly complex.

For all tasks, the datasets were split in a stratified manner into training, validation, and evaluation sets in a 70\%-15\%-15\% ratio, respectively, and features were standardized. Model training was performed with the Adam optimizer, using Cross Entropy as the non-regularizing loss function, as all tasks are classification problems. The primary performance metric was the F1-Score, chosen for its suitability in handling imbalanced datasets compared to accuracy. The KAN implementation and training were performed using the \texttt{PyTorch} \citep{PyTorch} and \texttt{pykan} \citep{KAN1,KAN2} frameworks.

During the feature selection phase, KANs with $k=3$, $G=5$, and $g_e=0.05$ were trained non-adaptively for 80 epochs. The grid search spanned the ranges $\left[\lambda_\text{min},\lambda_\text{max}\right] \times \left[\tau_\text{min},\tau_\text{max}\right] = \left[0.01, 0.1\right] \times \left[0.001, 0.01\right]$, with each range including 20 equidistant values. If the Pareto front was not a singleton, the model achieving the highest F1-Score with up to 10 features was selected. This choice is strict, as most state-of-the-art models employ at least 15 features for these datasets.

For the model selection phase, higher-order KANs with $k=4$ were used. Grid search was performed over $G \in \left\{8, 10, 12, 15, 20, 30, 40, 50\right\}$ and $g_e \in \left\{0.0, 0.05, \dots, 1.0\right\}$. Each model instance was trained adaptively for 200 epochs, with the grid updated every $10$ epochs until epoch $150$. For the symbolic fitting's cost function of Eq. \eqref{eq12}, $\alpha=0.05$ and $\beta=1.5$ were chosen to prioritize $R^2$ over complexity, except in cases of extremely high complexity where the exponential penalty dominates. If the resulting Pareto front contained more than one elements, the model with the highest average F1-Score between the regular and the symbolic representation was selected.

\subsection{Fault Detection}\label{sec4.1}

For the fault detection task, all data samples were categorized into two classes: normal (N) and faulty (F), with the latter encompassing all fault types. Fault detection is generally simpler than fault classification, as the model only needs to distinguish between normal and anomalous data. However, this task suffers from a significant imbalance in class representation, which presents a major challenge.

In both the CWRU and MaFaulDa datasets, the normal class is severely underrepresented. For the CWRU dataset, the normal class constitutes only 3.96\% of the dataset, while the other classes range from 11.88\% to 23.76\%. Similarly, in the MaFaulDa dataset, the normal class accounts for just 2.51\%, with the remaining classes spanning from 7.02\% to 17.07\%. When restructured for fault detection, the imbalance becomes even more pronounced, with the normal class constituting only 3.96\% versus 96.04\% for the faulty class in the CWRU dataset, and 2.51\% versus 97.49\% for the faulty class in the MaFaulDa dataset. This extreme imbalance means that even a trivial classifier which predicts all entries as faulty, would achieve a high accuracy (e.g., 97.49\% for MaFaulDa) while failing to provide any meaningful insights.

To address this imbalance, a balancing strategy is required. Among the common approaches are undersampling the dominant class or oversampling the minority class using techniques such as the Synthetic Minority Oversampling (SMOTE) \citep{SMOTE}. For this work, undersampling was adopted; specifically, 30 samples from each fault class in the CWRU augmented dataset and 60 samples from each fault class in the MaFaulDa augmented dataset were randomly selected. This adjustment reduced the imbalance to 12.28\% normal versus 87.72\% faulty for CWRU and 22.86\% normal versus 77.14\% faulty for MaFaulDa. Although still imbalanced, these distributions are far more manageable.

\begin{figure}[t]
	\centering
	\includegraphics[width=0.7\textwidth]{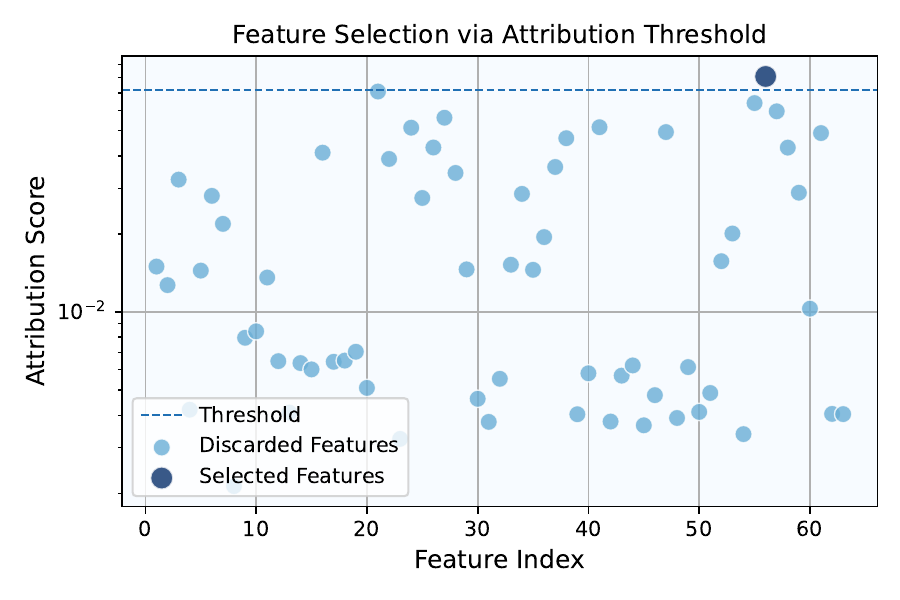}
	\caption{Feature attribution scores for all features within the feature library of CWRU. For the chosen threshold, $\tau = 7.16\cdot 10^{-2}$, only a single feature is selected.}
	\label{cwru_detect_feat_attrib}
\end{figure}

\begin{figure*}[t]
	\centering
	\includegraphics[width=\textwidth]{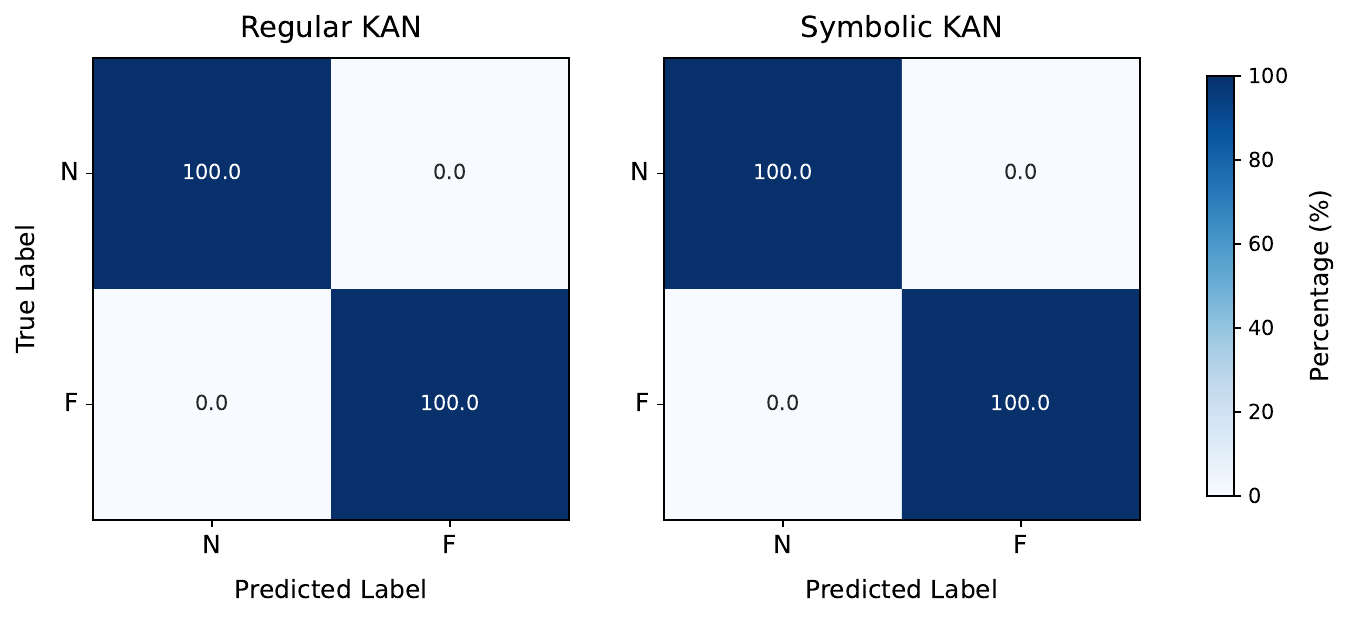}
	\caption{Confusion matrices for fault detection in the CWRU dataset by the regular (left) and symbolic (right) version of the trained KAN model.}
	\label{cwru_detect_CMs}
\end{figure*}

Following this preprocessing step, the framework's feature selection process, as detailed in Section \ref{sec2}, was applied to identify the most relevant features for fault detection. Regarding the CWRU dataset, the Pareto front resulting from the $\left(\lambda,\tau\right)$ grid search was a singleton, yielding $\lambda = 2.42\cdot 10^{-3}$ and $\tau = 7.16\cdot 10^{-2}$. These values resulted in the selection of a single feature, $x_{24}^2$ (the 25th percentile value for the drive-end signal, see Appendix \ref{appB}), as illustrated in Fig. \ref{cwru_detect_feat_attrib}. Proceeding to the model selection phase, the grid search using only this feature again produced a Pareto front with a single point, corresponding to $G=8$ and $g_e=0.0$. The combination of a single feature and a small, yet fully adaptive grid, suggests that fault detection in the CWRU dataset is relatively straightforward, so approaching the task with a complex model is neither necessary nor a good practice. This is further corroborated by the final evaluation of the chosen model, for both the regular and symbolic version of the KAN, as shown in the confusion matrices of Fig. \ref{cwru_detect_CMs}.

\begin{figure}[b]
	\centering
	\includegraphics[width=0.7\textwidth]{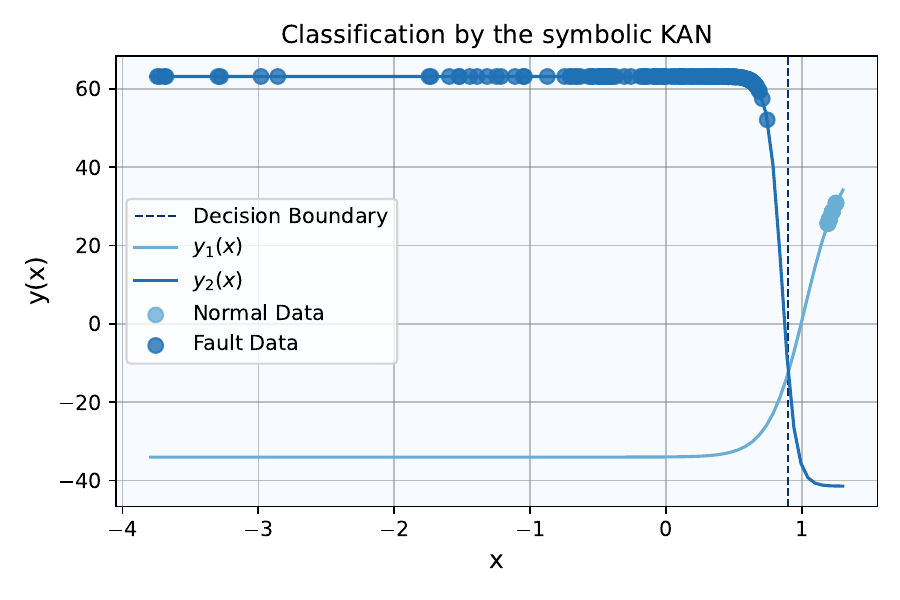}
	\caption{The symbolic KAN's output edges uniquely define class-exclusive regions separated by a decision boundary. The CWRU dataset's points are correctly classified in their respective categories.}
	\label{cwru_detect_symb_curves}
\end{figure}

The selection of a single feature offers the opportunity to highlight the importance of extracting symbolic representations for the trained KAN, as they constitute the model fully interpretable. In this case, the symbolic representations of the KAN's output edges are given by:

\begin{equation} y_1\left(x\right) = 42.62 - 76.65\sigma\left(7.69 - 7.51x\right), \label{eq13} \end{equation}

\noindent and

\begin{equation} y_2\left(x\right) = 10.85 - 52.3\tanh\left(10x - 8.52\right), \label{eq14} \end{equation}

\noindent where $x$ denotes the scaled feature and $\sigma\left(x\right) = 0.5\left[1 + \tanh\left(x/2\right)\right]$ is the sigmoid function. It is noted that all numbers have been rounded to the second digit. Using these analytical expressions, a sample is classified as Normal if $y_1\left(x\right) > y_2\left(x\right)$, and as a Fault if $y_1\left(x\right) < y_2\left(x\right)$. Eqs. \eqref{eq13} and \eqref{eq14} allow for the study of otherwise inaccessible (or hard to compute) properties of the classification problem, such as determining the decision boundary by solving $y_1\left(x\right) = y_2\left(x\right)$. Fig. \ref{cwru_detect_symb_curves} illustrates the two curves alongside all CWRU data points, color-coded by class. The decision boundary is also depicted, demonstrating that all of the dataset's samples are correctly classified using these symbolic expressions.

\begin{figure}[b]
	\centering
	\includegraphics[width=0.7\textwidth]{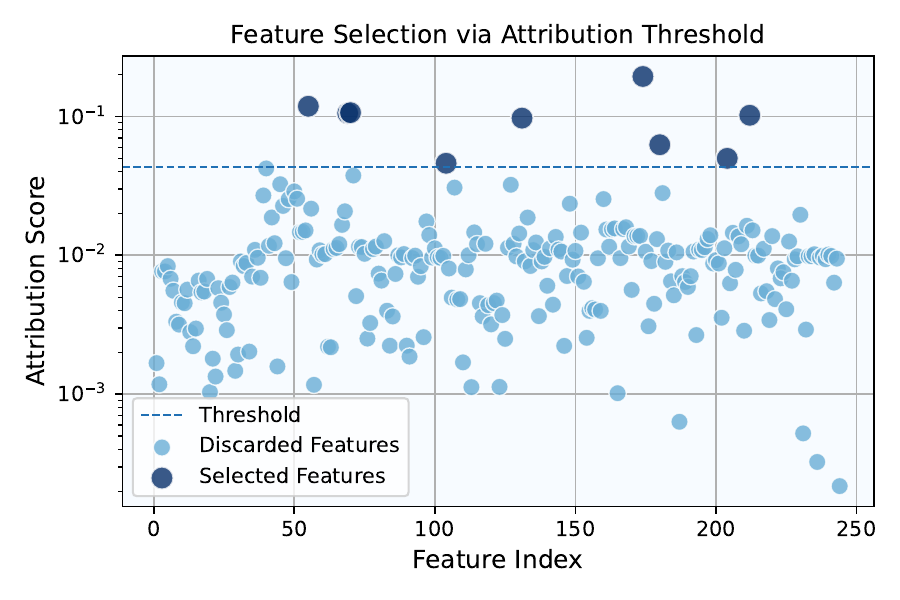}
	\caption{Feature attribution scores for all features within the feature library of MaFaulDa. For the chosen threshold, $\tau = 4.32\cdot 10^{-2}$, nine features are selected.}
	\label{maf_detect_feat_attrib}
\end{figure}

The same procedure was applied to the MaFaulDa dataset. In this case, the feature selection process resulted in a Pareto front with three candidate points. The corresponding $\left(\lambda,\tau\right)$ values, the associated F1-Scores for each point, and the number of features retained are presented in Table \ref{paretotab}. Although the configuration with the fewest features also exhibited the lowest performance, it still achieved a remarkably high F1-Score of 97.07\%. Notably, the four features selected in the lowest-performing case are a subset of the six features selected in the middle-performing case, which, in turn, are a subset of the nine features selected in the highest-performing case. This hierarchical relationship highlights the consistency of the framework. Following the selection rule of prioritizing the highest-performing configuration with no more than 10 features, the combination $\lambda = 2.9 \cdot 10^{-3}$ and $\tau = 4.32 \cdot 10^{-2}$ was chosen. This configuration retained nine features: $x_{6}^{2}$, $x_{15}^{4}$, $x_{30}^{4}$, $x_{4}^{5}$, $x_{30}^{5}$, $x_{31}^{5}$, $x_{11}^{6}$, $x_{2}^{7}$, and $x_{23}^{8}$, as illustrated in Fig. \ref{maf_detect_feat_attrib}.

\begin{table}[b]
	%\centering
	\caption{Combinations of $\lambda,\tau$ within the Pareto front and the corresponding model performance and number of retained features.}\label{paretotab}
	\begin{tabular}{@{}cccc@{}}
		\toprule
		$\lambda$ ($\cdot 10^{-3}$) & $\tau$ ($\cdot 10^{-2}$) & F1-Score & No. of features \\
		\midrule
		$1.00$            & $9.05$      & $97.07$\%		& $4$		\\
		$1.47$            & $6.68$      & $98.07$\%		& $6$	 	\\
		$2.90$            & $4.32$      & $100.00$\%		& $9$		\\
		\botrule
	\end{tabular}
\end{table}

\begin{figure*}[t]
	\centering
	\includegraphics[width=\textwidth]{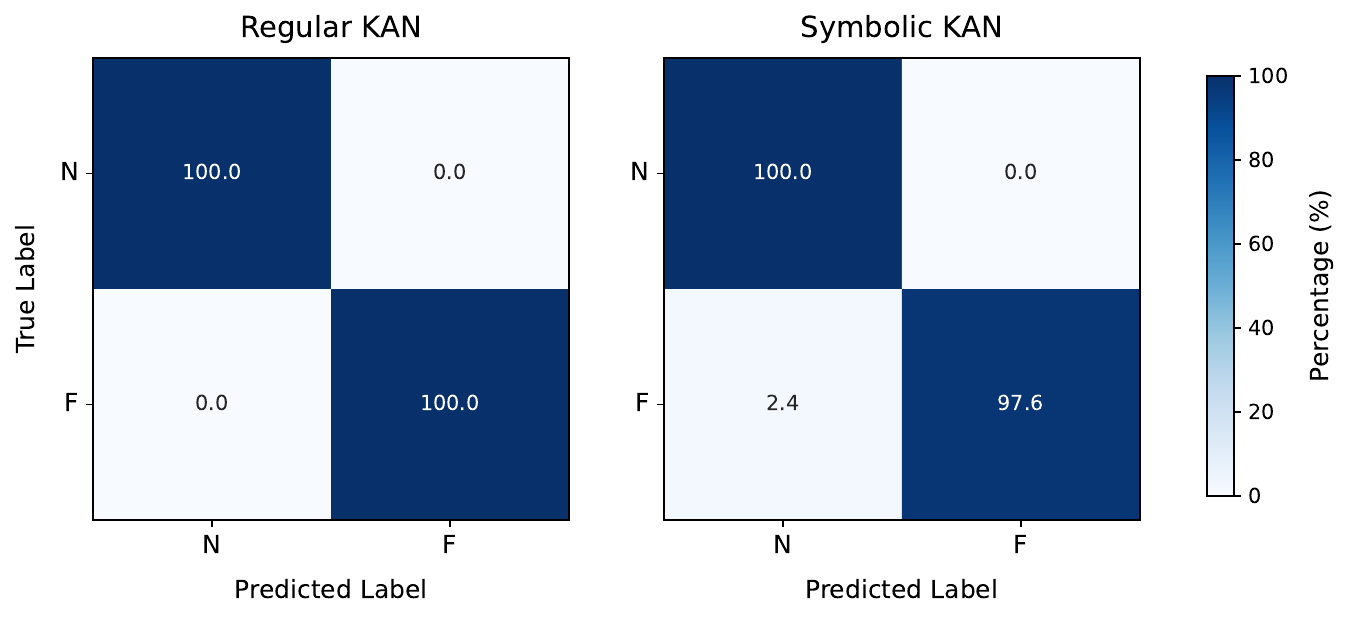}
	\caption{Confusion matrices for fault detection in the MaFaulDa dataset by the regular (left) and symbolic (right) version of the trained KAN model.}
	\label{maf_detect_CMs}
\end{figure*}

\begin{figure}[b]
	\centering
	\includegraphics[width=0.7\textwidth]{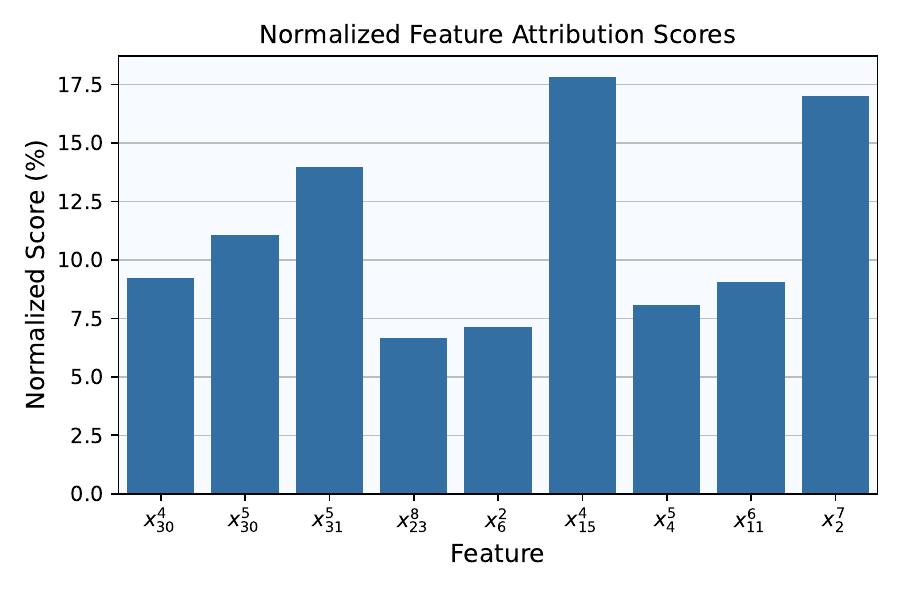}
	\caption{Final attribution scores (normalized) for the retained features of the trained model.}
	\label{maf_detect_feat_scores}
\end{figure}

For the model selection phase, using the nine selected features, the grid search over $G$ and $g_e$ produced a Pareto front with a single point, corresponding to $G=8$ and $g_e=0.05$, similar to the results for the CWRU dataset. The final trained model achieved a F1-Score of 100\% in its regular form and 98.08\% after symbolic fitting. This result is not unexpected, as symbolic fitting cannot always perfectly replicate the trained activation function using analytical expressions. Consequently, the trade-off between performance and interpretability must always be considered; however, in this case, the performance impact is minimal. The performance in terms of each KAN version's confusion matrices can be seen in Fig. \ref{maf_detect_CMs}. When it comes to the KAN's symbolic activation functions for this case, they are 9-dimensional, meaning that the visualization of their decision boundary - now corresponding to a 9-dimensional hypersurface - would require a 10-dimensional equivalent of Fig. \ref{cwru_detect_symb_curves}. Although such a depiction is impractical, these expressions remain computationally inexpensive and provide valuable insights into the model's predictions, for instance, by keeping most features constant and examining decision boundaries as one or two features vary. Notably, the presence of multiple features allows for an analysis of their final importance in the trained model's predictions. The normalized feature attribution scores, depicted in Fig. \ref{maf_detect_feat_scores}, illustrate the relative contribution of each selected feature.

\subsection{Fault Classification}\label{sec4.2}

\begin{figure*}[b]
	\centering
	\includegraphics[width=\textwidth]{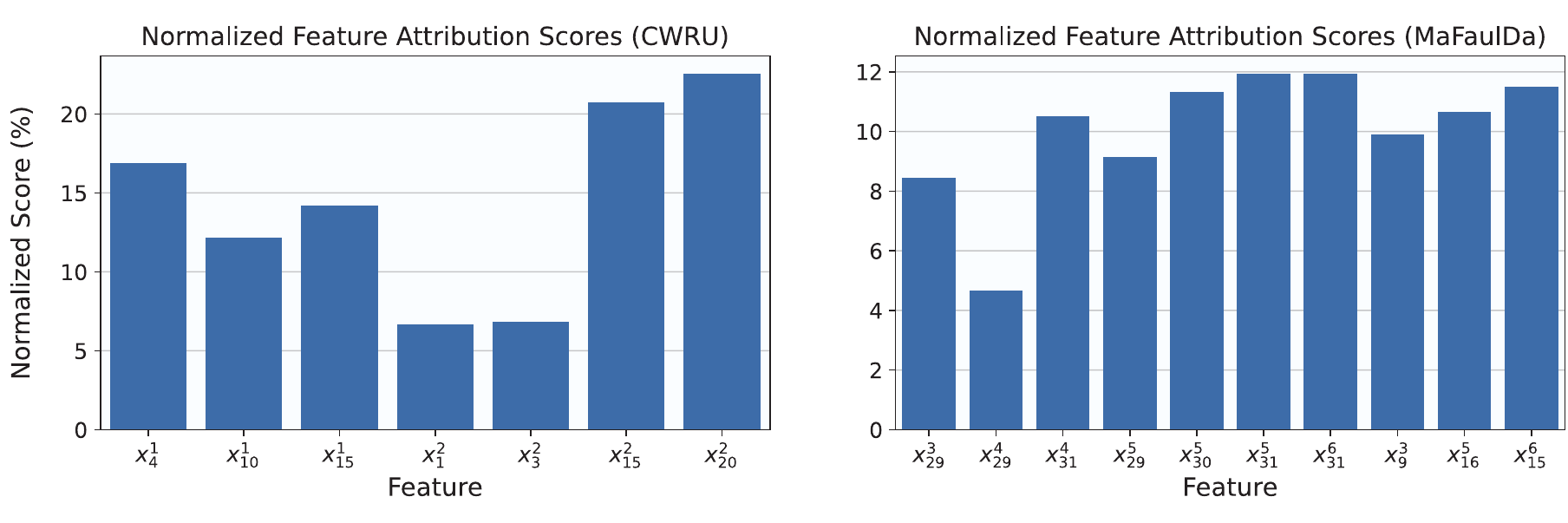}
	\caption{Normalized feature attribution scores after training the final model on the CWRU (left) and MaFaulDa (right) dataset.}
	\label{class_attribs}
\end{figure*}

\begin{figure*}[t]
	\centering
	\includegraphics[width=\textwidth]{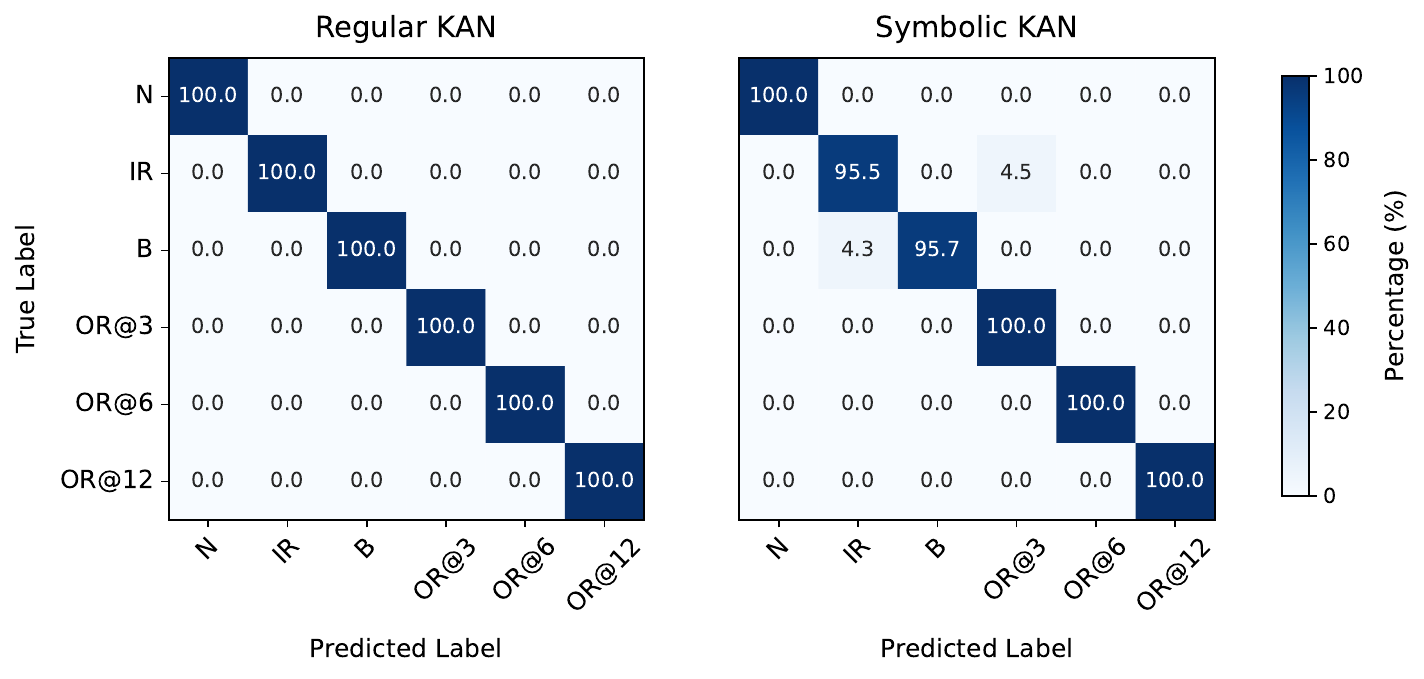}
	\caption{Confusion matrices for fault classification in the CWRU dataset by the regular (left) and symbolic (right) version of the trained KAN model.}
	\label{cwru_class_CMs}
\end{figure*}

Moving to fault classification, no oversampling or undersampling techniques were applied, despite the datasets being imbalanced as previously noted. For the CWRU dataset, the feature selection grid search yielded $\lambda = 10^{-3}$ and $\tau = 6.68 \cdot 10^{-2}$ as the optimal parameters, resulting in the selection of the following seven features: $x_{4}^{1}$, $x_{10}^{1}$, $x_{15}^{1}$, $x_{1}^{2}$, $x_{3}^{2}$, $x_{15}^{2}$, and $x_{20}^{2}$. During model selection, the corresponding grid search identified a model with $G = 12$ and $g_e = 0.5$ as achieving the highest average F1-Score between its regular and symbolic versions. The final feature importance scores for this model, evaluated on the evaluation set, are shown in the left plot of Fig. \ref{class_attribs}. The model in its regular form achieved a perfect F1-Score of 100\%, while the symbolic version slightly underperformed at 97.80\%. The corresponding confusion matrices for both versions are provided in Fig. \ref{cwru_class_CMs}.

\begin{figure*}[b]
	\centering
	\includegraphics[width=\textwidth]{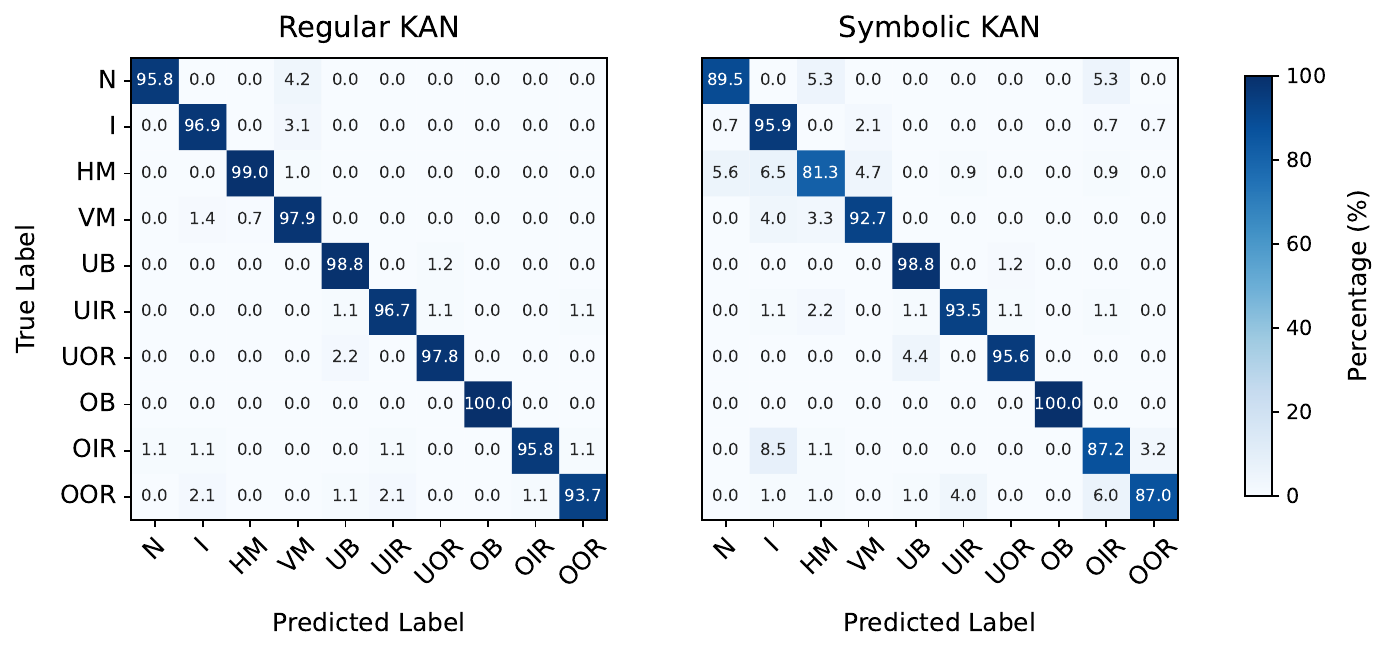}
	\caption{Confusion matrices for fault classification in the MaFaulDa dataset by the regular (left) and symbolic (right) version of the trained KAN model.}
	\label{maf_class_CMs}
\end{figure*}

Applying the same approach to the MaFaulDa dataset, the feature selection process yielded a large Pareto front, ranging from models that retained a single feature (with poor performance) to those retaining up to 70 features (achieving a validation set F1-Score of 99.89\%). The selected configuration, $\lambda = 7.63 \cdot 10^{-3}$ and $\tau = 3.37 \cdot 10^{-2}$, resulted in the retention of ten features: $x_{9}^{3}$, $x_{29}^{3}$, $x_{29}^{4}$, $x_{31}^{4}$, $x_{16}^{5}$, $x_{29}^{5}$, $x_{30}^{5}$, $x_{31}^{5}$, $x_{15}^{6}$, and $x_{31}^{6}$. The subsequent grid search for optimal model hyperparameters produced a smaller Pareto front with only two points. Among these, the configuration $G = 12$ and $g_e = 0.15$ was chosen and evaluated on the evaluation set. The selected model achieved a final F1-Score of 97.24\% in its regular form and 92.03\% in its symbolic form. The normalized feature attribution scores for the final model are shown in the right plot of Fig. \ref{class_attribs}, while the corresponding confusion matrices are depicted in Fig. \ref{maf_class_CMs}. %The symbolic expressions for both the CURW and MaFaulDa symbolic KANs are provided in the supplementary material.

It becomes evident that, unlike in the fault detection task, fault classification requires a greater number of features for each dataset to capture the details required for a more fine-grained classification than a binary one. For the CWRU dataset, the trained KAN once again achieved a perfect F1-Score in its regular form. However, for the MaFaulDa dataset, even the regular version of the KAN left some data points misclassified. This is not unexpected, as the MaFaulDa dataset includes not only bearing faults but also additional machinery faults. This, in fact, highlights the generalizability of the proposed framework, demonstrating its ability to perform well in more diverse scenarios beyond the narrower domain of bearing faults. It is worth noting that higher performance could have been achieved by increasing the number of selected features or adding more layers to the KAN. However, the focus of this study is not solely on achieving perfect scores but rather on promoting lightweight models with minimal parameters, favouring interpretability and ensuring suitability for deployment in resource-constrained environments.

\subsection{Severity Classification}\label{sec4.3}

Apart from fault detection and classification, both datasets allow for an additional type of investigation: the analysis of fault severity. For the CWRU dataset, all faults have three severity levels: 7 mils, 14 mils, and 21 mils. An exception to this are OR@12 faults, which only have two severity categories: 7 mils and 21 mils. For the MaFaulDa dataset, severity classification encompasses a broader range of categories across different fault types. Imbalance faults are categorized into seven severity levels, corresponding to imbalance loads ranging from 6 g to 35 g. Horizontal misalignment faults have four severity levels (0.5 mm, 1.0 mm, 1.5 mm, and 2.0 mm), while vertical misalignment faults have six severity levels (0.51 mm, 0.63 mm, 1.27 mm, 1.4 mm, 1.78 mm, and 1.9 mm). Additionally, the bearing faults in the MaFaulDa dataset are all classified into four severity levels, determined by the extra imbalance introduced to amplify their effects, ranging from 0 g to 20 g.

Given these structured severity levels, it is natural to approach severity analysis as a classification problem. To this end, the proposed framework was applied, following the established processes of feature selection, model selection, and model evaluation for each fault type's severity classification. Tables \ref{cwru_sev_tab} and \ref{maf_sev_tab} present the intermediate (i.e., feature selection and model selection outcomes) and final (i.e., F1-Scores for both the regular and symbolic KANs on the evaluation set) results for the CWRU and MaFaulDa datasets, respectively. %The symbolic expressions of the trained KANs, as well as the normalized feature attribution scores are provided in the supplementary material for completeness.

\begin{table*}[b]
	\centering
	\caption{Results for severity classification on the different fault types of the CWRU dataset. In the F1-Score column both the regular - symbolic versions are reported.}\label{cwru_sev_tab}
	\begin{tabular}{@{}clccc@{}}
		\toprule
		Fault & Features & $G$ & $g_e$ & F1-Score \\
		\midrule
		IR    	&  [$x_{21}^{2}$]   &  $8$		&  $0.00$   &  $95.45$\%   -   $100.00$\%    \\ \rule{0pt}{1.2em}
		B    	&  [$x_{15}^{1}$, $x_{28}^{1}$, $x_{27}^{2}$]    &  $10$		&  $0.40$   &  $100.00$\%   -   $95.45$\%    \\ \rule{0pt}{1.2em}
		OR@3    &  [$x_{28}^{1}$]    &  $8$		&  $0.00$   &  $100.00$\%   -   $100.00$\%    \\ \rule{0pt}{1.2em}
		OR@6    &  [$x_{21}^{1}$, $x_{15}^{2}$]    &  $8$		&  $0.00$   &  $100.00$\%  -   $100.00$\%    \\ \rule{0pt}{1.2em}
		OR@12   &  [$x_{25}^{1}$]    &  $8$		&  $0.50$   &  $100.00$\%   -   $100.00$\%    \\ 
		\botrule
	\end{tabular}
\end{table*}

\begin{table*}[t]
	\centering
	\caption{Results for severity classification on the different fault types of the MaFaulDa dataset. In the F1-Score column both the regular - symbolic versions are reported.}\label{maf_sev_tab}
	\begin{tabular}{@{}clccc@{}}
		\toprule
		Fault & Features & $G$ & $g_e$ & F1-Score \\
		\midrule
		I    	&  [$x_{23}^{1}$, $x_{1}^{3}$, $x_{24}^{3}$, $x_{25}^{3}$, $x_{1}^{4}$, $x_{11}^{4}$, $x_{18}^{5}$, $x_{31}^{5}$, $x_{18}^{6}$, $x_{15}^{8}$]    &  $10$	&  $0.05$  &  $95.71$\%    -   $91.35$\%    \\ \rule{0pt}{1.2em}
		HM    	&  [$x_{2}^{3}$, $x_{3}^{3}$, $x_{24}^{3}$, $x_{25}^{3}$, $x_{6}^{4}$, $x_{9}^{4}$, $x_{31}^{6}$, $x_{2}^{8}$]   &  $10$		&  $0.10$   &  $100.00$\%   -   $100.00$\%    \\ \rule{0pt}{1.2em}
		VM    	&  [$x_{23}^{1}$, $x_{26}^{1}$, $x_{22}^{3}$, $x_{24}^{3}$, $x_{6}^{4}$, $x_{18}^{5}$, $x_{19}^{5}$, $x_{30}^{5}$, $x_{18}^{6}$,  $x_{6}^{8}$]    &  $20$		&  $0.10$   &  $86.67$\%   -   $87.40$\%    \\ \rule{0pt}{1.2em}
		UB    	&  [$x_{9}^{2}$, $x_{1}^{3}$, $x_{1}^{4}$, $x_{1}^{5}$, $x_{1}^{7}$, $x_{30}^{7}$, $x_{6}^{8}$]    &  $8$		&  $0.20$   &  $98.89$\%   -   $97.79$\%    \\ \rule{0pt}{1.2em}
		UIR   	&  [$x_{3}^{2}$, $x_{11}^{3}$, $x_{25}^{3}$, $x_{29}^{3}$, $x_{1}^{4}$, $x_{4}^{4}$, $x_{1}^{5}$, $x_{30}^{5}$, $x_{4}^{7}$]    &  $8$		&  $0.05$   &  $100.00$\%   -   $100.00$\%    \\ \rule{0pt}{1.2em}
		UOR    	&  [$x_{23}^{1}$, $x_{2}^{2}$, $x_{4}^{3}$, $x_{8}^{3}$, $x_{1}^{4}$, $x_{4}^{4}$, $x_{30}^{6}$]    &  $10$		&  $0.00$   &  $100.00$\%   -   $95.54$\%    \\ \rule{0pt}{1.2em}
		OB    	&  [$x_{1}^{2}$, $x_{1}^{3}$, $x_{1}^{4}$, $x_{1}^{5}$, $x_{1}^{6}$, $x_{3}^{6}$, $x_{1}^{7}$, $x_{9}^{7}$]    &  $20$		&  $0.40$   &  $96.48$\%   -  $94.29$\%    \\ \rule{0pt}{1.2em}
		OIR   	&  [$x_{2}^{3}$, $x_{1}^{4}$, $x_{15}^{4}$, $x_{4}^{6}$, $x_{15}^{6}$, $x_{31}^{7}$]    &  $8$		&  $0.15$   &  $100.00$\%   -   $97.82$\%    \\ \rule{0pt}{1.2em}
		OOR   	&  [$x_{1}^{3}$, $x_{2}^{3}$, $x_{4}^{3}$, $x_{8}^{3}$, $x_{1}^{4}$, $x_{9}^{4}$, $x_{1}^{7}$, $x_{4}^{7}$]    &  $8$		&  $0.20$   &  $98.91$\%   -   $98.91$\%    \\
		\botrule
	\end{tabular}
\end{table*}

Based on these results, previous observations regarding the relative simplicity of the CWRU dataset are reaffirmed: once again, the severity of each fault type can be classified with perfect accuracy using a very small number of features (no more than three), with only a single misclassification occurring in the case of IR faults. Notably, in this instance, the symbolic version of the KAN outperforms the regular version, achieving perfect classification accuracy. Another indicator of the CWRU dataset's simplicity is that the model instances consistently utilize the smallest permitted grid size ($G = 8$), with only one exception requiring $G = 10$. In contrast, the MaFaulDa dataset demonstrates greater complexity. Larger grid sizes, such as $G = 20$, are employed in several cases, and the number of utilized features is generally higher, often reaching the maximum of ten to achieve the reported results. Nonetheless, with the exception of vertical misalignment faults, all severities across all fault categories achieve F1-Scores exceeding 95\% for the regular KAN, underscoring the effectiveness of the proposed framework even in more diverse and challenging scenarios.

Interestingly, the patterns observed for feature prevalence in fault detection and classification tasks do not carry over to severity classification. Specifically, while time-domain and frequency-domain features dominated the fault detection and classification tasks for CWRU, and wavelet-based features were dominant for MaFaulDa, the opposite trend is observed in the severity classification task. Approximately two-thirds of the selected features for severity classification in CWRU and MaFaulDa are wavelet-based features and time- or frequency-domain features, respectively. This observation highlights the distinction between fault identification/classification and severity quantification. From a feature selection perspective, it confirms that no single feature set is universally suitable for all tasks, emphasizing the value of the proposed framework for automatic feature selection from a diversified feature library.

\section{Discussion and Conclusion}\label{sec5}

In the present study, a novel framework leveraging KANs was developed and applied to bearing fault detection \& classification, as well as fault severity classification tasks. The framework utilizes an attribution scoring mechanism coupled with a grid-search-based multi-objective optimization procedure for automatic feature selection from an extensive feature library and hyperparameter tuning. By design, it emphasizes lightweight models with interpretable outputs, which is achieved by training shallow KANs and subsequently replacing their activation functions with analytical expressions drawn from a symbolic library. This approach prioritizes not only high performance, but also deployability and explainability, both of which are essential for real-world applications.

The framework was validated using two widely recognized datasets, CWRU and MaFaulDa, which were processed and augmented to enable feature extraction for the construction of a feature library. The experimental results demonstrated the framework's effectiveness across all tasks. For fault detection, it achieved perfect performance on both datasets, highlighting its capability to distinguish between normal and faulty conditions even under significant class imbalance. In fault classification, the CWRU dataset proved relatively straightforward, with the framework again achieving perfect F1-Scores. In contrast, the MaFaulDa dataset required more features and larger grids to handle its increased complexity, yet the regular version of the trained KAN still achieved a 97.24\% F1-Score on the evaluation set. For severity classification, the framework accurately identified fault severity levels with high F1-Scores in the regular KAN model instances: 100\% for four out of five fault types of the CWRU dataset and greater than 95\% for all fault types in the MaFaulDa dataset, with the exception of VM faults. These results showcase the framework's adaptability across diverse fault types and severity categories. Finally, while the symbolic versions of the models occasionally exhibited slightly reduced performance compared to their regular version, they proved invaluable in terms of the interpretability they offer.

\begin{figure}[b]
	\centering
	\includegraphics[width=0.65\textwidth]{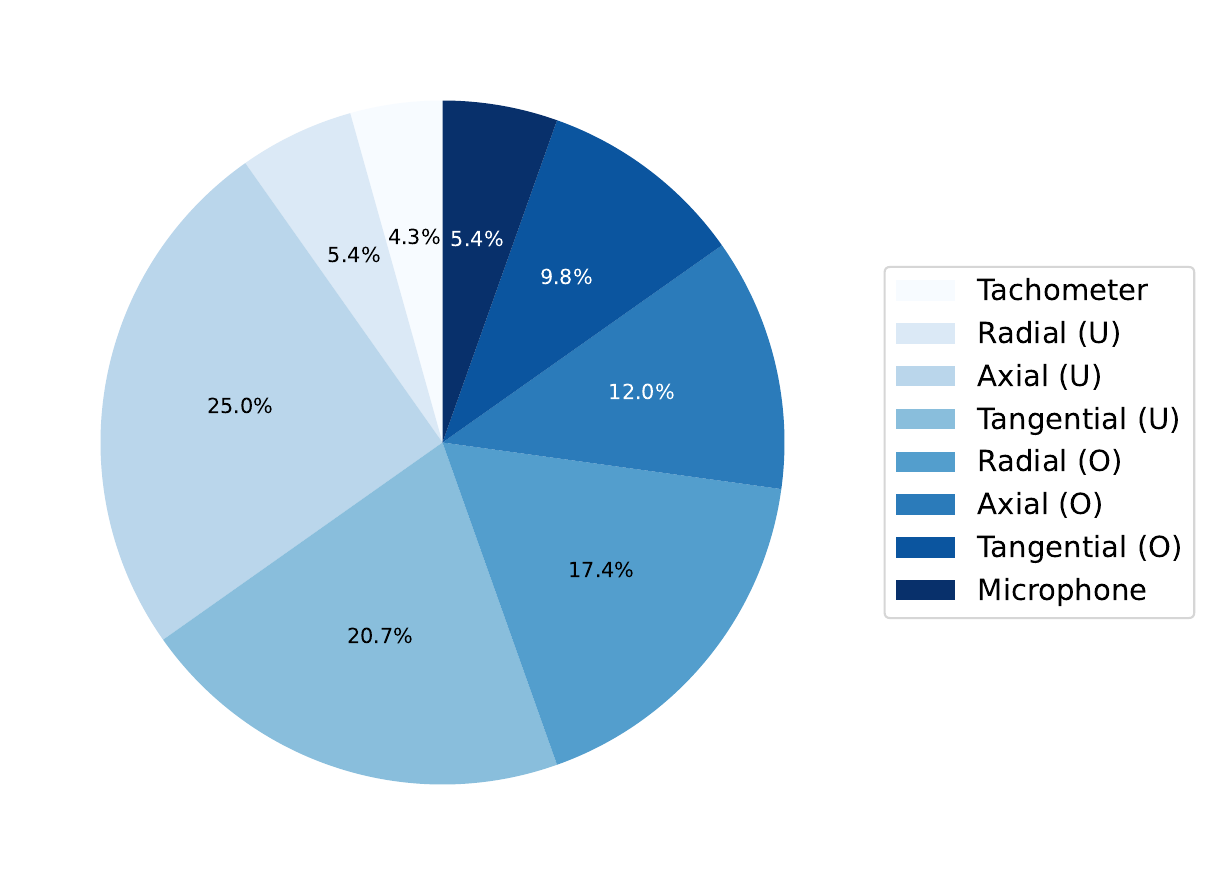}
	\caption{Pie chart of the frequency with which each of the eight signals available in the MaFaulDa dataset appeared in a task's selected features.}
	\label{pies}
\end{figure}

Explainability is a critical aspect of this framework, which is why shallow KANs with a limited number of features were prioritized. Beyond solving the task at hand, the framework can provide significant insights, such as identifying the most relevant signal types for each task or dataset. For instance, Fig. \ref{pies} offers a breakdown of the frequency of each signal's contribution to the feature library for the MaFaulDa dataset. Across all seven experiments conducted for the CWRU dataset (fault detection, fault classification, and five severity analyses), the drive-end and fan-end signals were equally represented among the selected features, with a perfect 50\%-50\% ratio. Conversely, for the MaFaulDa dataset, the tachometer and microphone signals were consistently underrepresented in the selected features, underscoring their limited relevance to the studied tasks. In contrast, signals from the underhang bearing (denoted by U) accounted for approximately 50\% of the selected features across all tasks. Such observations, along with the findings on the dominant feature types for specific tasks mentioned in Section \ref{sec4}, emphasize the value of explainability.

Combining all of the above, a practical implementation of this framework in a real-world scenario could focus on identifying or classifying bearing faults - or any type of machinery fault, given the framework's generalizability - in industries such as manufacturing or the energy sector. Using historical data, the framework could first identify the optimal set of sensors to install on the machinery, reducing costs by focusing only on the most informative signals. Once the sensors are installed, signal data could be used to construct feature libraries based on domain knowledge, while the framework automates feature selection for the specific tasks at hand. The resulting lightweight models can be deployed online for real-time inference through MLOps platforms like MLflow \citep{mlflow}, with inference being rapid due to the models' minimal computational requirements. Periodic retraining of the models could also be seamlessly integrated into the same platform, with short training times owing to the models' small number of parameters. Furthermore, in the event of other types of faults or issues arising within the machinery, the framework's versatility would allow it to be extended to address and analyze these problems as well.

Beyond engineering applications, the framework also holds significant potential for scientific tasks, particularly in the domain of symbolic regression. Building on the groundwork of \cite{KAN2}, this framework offers an alternative to traditional symbolic regression methods, avoiding the computational overhead typically associated with genetic algorithms. For instance, in scientific problems where the underlying equation describing the data is not known, the cost function defined in Eq. \eqref{eq12} could be utilized to generate multiple symbolic expressions per run by varying the parameters $\alpha$ and $\beta$, effectively creating a grid-search process similar to those detailed in this work. From the resulting expressions, the optimal one could be selected using a defined metric or based on domain-specific knowledge, such as dimensional constraints.

\newpage

\begin{appendices}
	
	\section{Symbolic Functions Library} \label{appA}
	
	This Appendix contains the symbolic library, shown in Table \ref{symblib}, which consists of a predefined set of univariate functions for replacing the trained activation functions of KANs with interpretable symbolic representations. The table also depicts the assigned complexities and analytical forms of the functions.
	
	\begin{table*}[h]
		\centering
		\caption{Library of symbolic functions and assigned complexities.}\label{symblib}%
		\begin{tabular}{@{}ccc@{}}
			\toprule
			Symbolic Function & $f\left(x\right)$ & $C$\\
			\midrule
			Zero                   & $0$                     & 1 \\
			Linear                 & $x$                     & 1 \\
			Exponential            & $\exp(x)$               & 2 \\
			Logarithmic            & $\ln(x)$                & 2 \\
			Absolute Value         & $|x|$                   & 2 \\
			Sine                   & $\sin(x)$               & 2 \\
			Cosine                 & $\cos(x)$               & 2 \\
			Tangent Hyperbolic     & $\tanh(x)$              & 2 \\
			Sign Function          & $\text{sgn}(x)$         & 2 \\
			Arctangent             & $\arctan(x)$            & 2 \\
			Hyperbolic Cosine      & $\cosh(x)$              & 2 \\
			Square Root            & $\sqrt{x}$              & 3 \\
			Quadratic              & $x^2$                   & 3 \\
			Cubic                  & $x^3$                   & 3 \\
			Quartic                & $x^4$                   & 3 \\
			Quintic                & $x^5$                   & 3 \\
			Reciprocal             & $\frac{1}{x}$           & 3 \\
			Reciprocal Square      & $\frac{1}{x^2}$         & 5 \\
			Reciprocal Cubic       & $\frac{1}{x^3}$         & 5 \\
			Reciprocal Quartic     & $\frac{1}{x^4}$         & 5 \\
			Reciprocal Quintic     & $\frac{1}{x^5}$         & 5 \\
			Reciprocal Square Root & $\frac{1}{\sqrt{x}}$    & 5 \\
			Gaussian               & $\exp(-x^2)$            & 6 \\
			Sigmoid                & $\frac{1}{1+\exp(-x)}$  & 6 \\
			\botrule
		\end{tabular}
	\end{table*}

	\section{Feature Library and Notation} \label{appB}
	
	This Appendix presents the features included in the feature libraries constructed for each dataset, along with the correspondence between their $x_{n}^{i}$ notation and their actual feature names. This notation is used for brevity in figures and symbolic expressions, where the subscript $n$ corresponds to a specific feature index as shown in the Table \ref{featlib}, and the superscript $i$ denotes a dataset's signal. These features were processed as described in Section \ref{sec4}. For the CWRU dataset, $x_{n}^1$ and $x_{n}^2$ represent the fan-end and drive-end signals, respectively. For the MaFaulDa dataset, $x_{n}^{i}$ corresponds to: the tachometer signal for $i=1$; the radial, axial, and tangential accelerometer signals on the underhang bearing for $i = 2, 3, 4$, respectively; the radial, axial, and tangential accelerometer signals on the overhang bearing for $i = 5, 6, 7$, respectively; the microphone signal for $i=8$.
	
	\begin{table*}[h]
	\centering
	\caption{Feature notation, name and domain for each feature of the feature library.}\label{featlib}
	\begin{tabular}{@{}ccc@{}}
		\toprule
		Notation & Name & Domain \\
		\midrule
		$x_1^i$                   & Magnitude at fundamental frequency      & Frequency				\\
		$x_2^i$                   & Magnitude at second harmonic            & Frequency				\\
		$x_3^i$                   & Magnitude at third harmonic             & Frequency				\\
		$x_4^i$                   & Spectral Skewness                       & Frequency				\\
		$x_5^i$                   & Spectral Kurtosis                       & Frequency				\\
		$x_6^i$                   & Statistical Mean                        & Time				\\
		$x_7^i$                   & Statistical Variance                    & Time				\\
		$x_8^i$                   & Statistical Kurtosis                    & Time				\\
		$x_9^i$                   & Statistical Skewness                    & Time				\\
		$x_{10}^i$                & Statistical RMS                         & Time				\\
		$x_{11}^i$                & Shape Factor                            & Time				\\
		$x_{12}^i$                & Crest Factor                            & Time				\\
		$x_{13}^i$                & Impulse Factor                          & Time				\\
		$x_{14}^i$                & Margin Factor                           & Time				\\
		$x_{15}^i$                & Shannon Entropy                         & Time				\\
		$x_{16}^i$                & Histogram Upper Bound                   & Time				\\
		$x_{17}^i$                & Histogram Lower Bound                   & Time				\\
		$x_{18}^i$                & Wavelet Mean                            & Time-Frequency				\\
		$x_{19}^i$                & Wavelet Median                          & Time-Frequency				\\
		$x_{20}^i$                & Wavelet RMS                             & Time-Frequency				\\
		$x_{21}^i$                & Wavelet Standard Deviation              & Time-Frequency				\\
		$x_{22}^i$                & Wavelet Variance                        & Time-Frequency				\\
		$x_{23}^i$                & 5th percentile value           	  		& Time-Frequency				\\
		$x_{24}^i$                & 25th percentile value       	        & Time-Frequency				\\
		$x_{25}^i$                & 75th percentile value         		    & Time-Frequency				\\
		$x_{26}^i$                & 95th percentile value          		    & Time-Frequency				\\
		$x_{27}^i$                & Mean Crossings                          & Time-Frequency				\\
		$x_{28}^i$                & Zero Crossings                          & Time-Frequency				\\
		$x_{29}^i$                & Wavelet Shannon Entropy                 & Time-Frequency				\\
		$x_{30}^i$                & Wavelet Skewness                        & Time-Frequency				\\
		$x_{31}^i$                & Wavelet Kurtosis                        & Time-Frequency				\\
		\botrule
	\end{tabular}
	\end{table*}
	
\end{appendices}
\newpage
\bigskip

\backmatter

%\bmhead{Acknowledgements} S. R. and M. P. would like to thank A. Goumas for his invaluable assistance in creating some figures for this paper.

\bmhead{Author contributions} Conceptualization: Spyros Rigas; Methodology: Spyros Rigas; Software: Spyros Rigas, Michalis Papachristou; Data curation: Ioannis Sotiropoulos; Writing - original draft preparation: Spyros Rigas, Michalis Papachristou, Georgios Alexandridis; Supervision: Georgios Alexandridis.

\bmhead{Funding} No funding was received for conducting this study.

\bmhead{Data availability} The original MaFaulDa time-series dataset is available at \url{https://www02.smt.ufrj.br/~offshore/mfs}. The original CWRU time-series dataset is available at \url{https://engineering.case.edu/bearingdatacenter}. The extracted feature datasets and code used to produce the results of all experiments presented herein are available at \url{https://github.com/srigas/KAN_Fault_Diagnosis}.

\section*{Declarations}

\bmhead{Conflict of interest} The authors have no competing interests to declare that are relevant to the content of this article.

\bibliography{sn-bibliography}% common bib file

\end{document}